\documentclass[sigconf]{acmart}
\AtBeginDocument{%
  }

\setcopyright{acmlicensed}
\copyrightyear{2018}
\acmYear{2018}
\acmDOI{XXXXXXX.XXXXXXX}
\acmConference[Conference acronym 'XX]{Make sure to enter the correct
  conference title from your rights confirmation email}{June 03--05,
  2018}{Woodstock, NY}
\acmISBN{978-1-4503-XXXX-X/2018/06}

\usepackage{booktabs,multirow,graphicx,array}

\begin{document}

\title{ChainClaw: A Layered Agent Framework for Reliable On-Chain Execution}
\author{%
  {\small
  Jiacheng Wei\textsuperscript{1,2},
  Zhaoxin Fan\textsuperscript{1,2,*},
  Xin Wen\textsuperscript{2},
  Yuqin Lan\textsuperscript{2},
  Dongrun Li\textsuperscript{2},
  Wenjun Wu\textsuperscript{2},
  Faguo Wu\textsuperscript{1,2,4,*},
  Xiao Zhang\textsuperscript{2,3,4,*}
  }
}

\affiliation[obeypunctuation=true]{%
  \institution{%
    \textsuperscript{1}Beijing Advanced Innovation Center for Future
    Blockchain and Privacy Computing, Beihang University, Beijing, China\\
    \textsuperscript{2}School of Artificial Intelligence,
    Beihang University, Beijing, China\\
    \textsuperscript{3}School of Mathematical Sciences,
    Beihang University, Beijing, China\\
    \textsuperscript{4}Zhongguancun Laboratory,
    Beihang University, Beijing, China
  }
  \city{}
  \country{}
}

\email{jakiewei258@gmail.com}
\thanks{%
  \textsuperscript{*}Corresponding authors:
  Zhaoxin Fan, Faguo Wu, and Xiao Zhang.
}









\begin{abstract}
General-purpose large language model agents have achieved strong performance on tool-augmented tasks, yet they rely on assumptions break down in blockchain environments. On-chain execution is stateful, adversarial, and economically irreversible, exposing three fundamental gaps: Reactivity, Irreversibility, and Observability. We propose ChainClaw, a blockchain-native agent framework built on OpenClaw, that addresses all three gaps through a layered architecture comprising an event-driven orchestration layer, a simulation-based safety intelligence layer, and an on-chain monitoring runtime layer, unified by a cross-layer memory subsystem. ChainClaw closes the Reactivity gap via event ingestion and simulation feedback, the Irreversibility gap via a pre-execution safety pipeline with transaction simulation and action guard, and the Observability gap via an on-chain read adapter and transaction monitor. We evaluate ChainClaw on a purpose-built benchmark covering seven tasks across four categories and five dimensions. ChainClaw consistently outperforms representative baselines on both safety and task completion. 
\end{abstract}


\keywords{Do, Not, Use, This, Code, Put, the, Correct, Terms, for,
  Your, Paper}

\received{20 February 2007}
\received[revised]{12 March 2009}
\received[accepted]{5 June 2009}

\maketitle
\renewcommand{\shortauthors}{Wei et al.}
\section{Introduction}
Recent advances in tool-augmented large language model agents have enabled general-purpose systems to decompose user intents, invoke external tools, and complete complex interactive tasks~\cite{yao2022react, qin2024toolllm, schick2023toolformer}. Frameworks such as OpenClaw~\cite{openclaw2025} exemplify this trend by providing a modular agent runtime that integrates planning, 

\begin{figure}[H]
\centering
\includegraphics[scale=0.38]{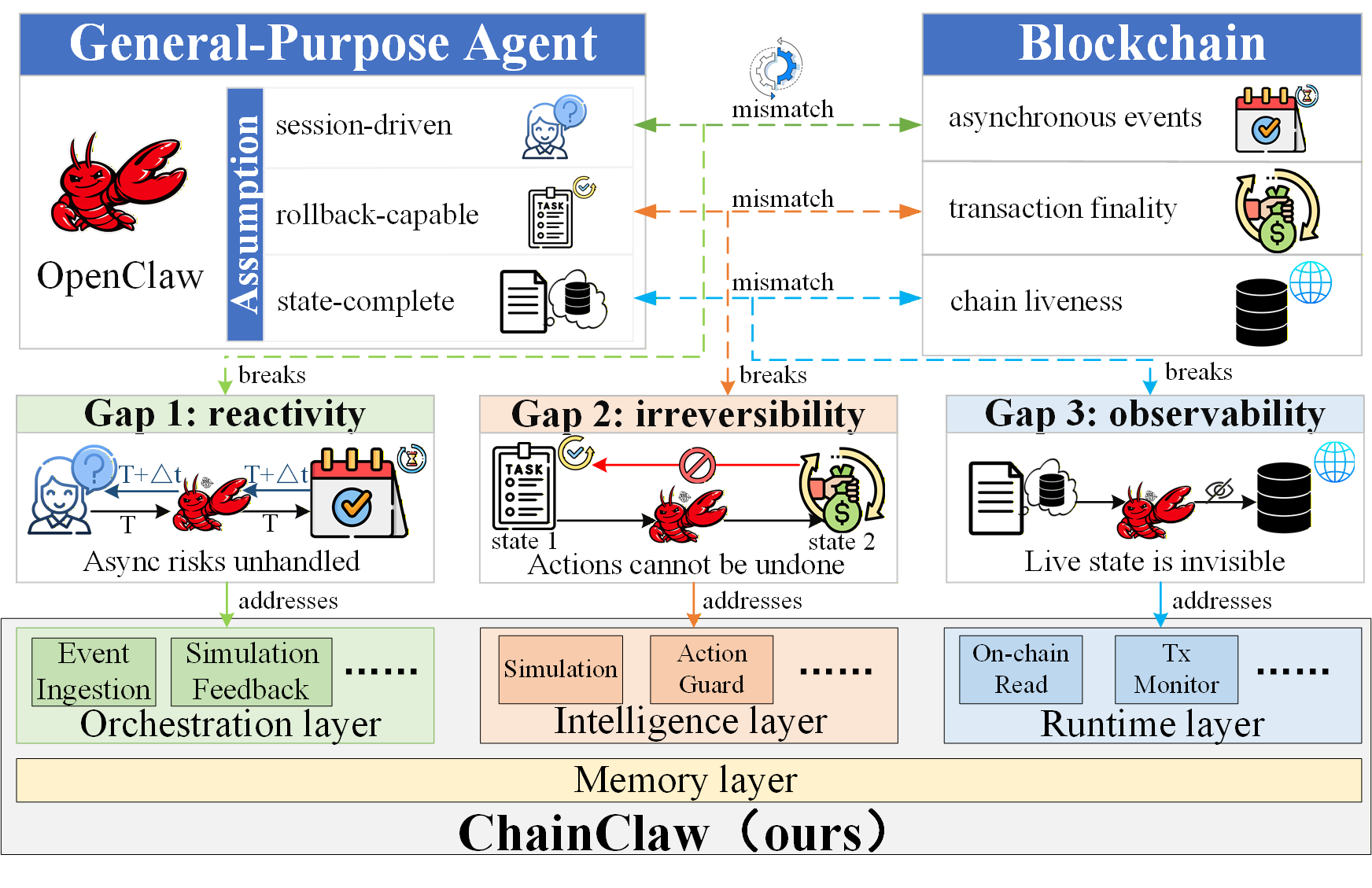}
\caption{General-purpose agents fail on blockchain due to reactivity, irreversibility, and observability gaps. ChainClaw addresses them through event-driven orchestration  layer, simulation-based safety intelligence layer, and on-chain monitoring runtime layer.}
\label{motivation}
\vspace{-1em}
\end{figure}

\noindent memory, tool execution, and feedback into a unified control loop~\cite{wang2024survey,xi2025rise}. Such systems have been widely adopted because they offer a flexible abstraction for connecting language models with heterogeneous tools~ \cite{shen2023hugginggpt, patil2024gorilla}, allowing agents to operate beyond pure text generation. In many conventional settings, tool calls are treated as recoverable actions~\cite{shinn2023reflexion, madaan2023self}, working well for tasks such as information retrieval, file manipulation, and software automation where failures are often correctable after execution~\cite{jimenez2024swe, yang2024swe, zhou2024webarena}.

However, blockchain environments expose a fundamentally different execution setting. An on-chain agent does not merely call external tools; it interacts with a stateful, adversarial, and economically sensitive runtime~\cite{zhou2023sok,babel2023clockwork,alqithami2026autonomous}. Once a transaction is signed and broadcast, it may consume gas, alter global state, grant token permissions, or transfer assets in ways that cannot be undone~\cite{liu2025blockchain,sun2024gptscan,hu2025walletprobe}. Moreover, the correctness of an on-chain action depends on live external state,  which are not contained in the agent's internal context~\cite{xi2024pomabuster,zhang2025following}. These properties conflict with three implicit assumptions  when general-purpose frameworks such as OpenClaw are deployed on-chain. First, the assumption that interaction is primarily session-driven is broken by asynchronous on-chain events and delayed transaction outcomes, leading to a Reactivity gap. Second, the assumption that execution is recoverable is broken by irreversible state-changing transactions, leading to an Irreversibility gap. Third, the assumption that the agent context contains sufficient task state is broken by live and external blockchain state, leading to an Observability gap. These gaps suggest that adapting a general-purpose agent framework to blockchain requires a dedicated on-chain agent architecture.

To this end, we propose ChainClaw, a unified on-chain agent framework built on top of OpenClaw, extending its modular runtime with blockchain-native components designed to close all three gaps. ChainClaw adopts a layered architecture that separates high-level task orchestration, agent intelligence, blockchain runtime execution, and chain-state access, supported by a cross-layer memory subsystem. The orchestration layer normalizes user requests and asynchronous on-chain events into executable tasks. The intelligence layer performs reasoning, knowledge retrieval, simulation, and policy checking. The runtime layer bridges abstract agent decisions with concrete blockchain operations, including on-chain reads, transaction execution, monitoring, chain interfaces, and key-vault-based signing. The chain network layer exposes profiles, live state, transaction histories, and indexed events. This organization enables ChainClaw to maintain continuity across sessions, ground decisions in live state, and safely control transaction execution.

Built on this architecture, ChainClaw addresses each identified gap through dedicated modules. For Reactivity, it adds event ingestion and simulation feedback, allowing asynchronous blockchain events to trigger reasoning and replanning beyond the user session. For Irreversibility, it introduces a pre-execution safety pipeline consisting of transaction simulation and an action guard. Before signing or broadcasting a transaction, ChainClaw simulates the intended operation, checks potential state changes, detects unsafe asset movements or abnormal approvals, and blocks actions that violate safety constraints. For Observability, it incorporates an on-chain read adapter and a transaction monitor, enabling the agent to ground its reasoning in live chain state rather than stale memory or prompt context. In addition, the chain interface layer and key vault further ensure backend portability and secure signing across heterogeneous blockchain environments.

We evaluate ChainClaw on a purpose-built benchmark for on-chain agent execution, where it consistently outperforms all baselines across safety-critical and event-driven scenarios. The benchmark measures five dimensions: completion, correctness, safety, robustness, and efficiency. It contains seven tasks across four categories: query tasks, including ETH balance and token allowance queries; single-step execution tasks, including ETH transfer and token approval; multi-step execution tasks, including approve-then-swap workflows; and safety-critical abnormal tasks, including malicious asset-transfer attempts and passive event-triggered responses. We compare ChainClaw against representative agent baselines, including ReAct, LangChain, and OpenClaw. We further conduct a user study in which participants interact with the system and evaluate it using a unified scoring form. The user feedback confirms that ChainClaw provides effective support for on-chain task execution, particularly in improving perceived safety, task clarity, and confidence before transaction execution. Overall, our results demonstrate that ChainClaw is a practical and necessary step toward reliable blockchain-native agents.

Our contributions can be summarized as,
\vspace{-2em}

\begin{itemize}
    \item We identify three core challenges in deploying general-purpose agents on blockchain, namely the Reactivity, Irreversibility, and Observability gaps, which together motivate the need for a dedicated on-chain agent architecture.
    \item We propose ChainClaw, comprising an event-driven orchestration layer, a simulation-based safety intelligence layer, and an on-chain monitoring runtime layer, unified by a cross-layer memory subsystem, which collectively close all three identified gaps.
    \item  We design a benchmark covering seven tasks across four categories and five evaluation dimensions, and  ChainClaw consistently outperforms representative baselines on safety and task completion.
\end{itemize}

\section{Related Work}

\textit{LLM Agent Frameworks.}
Large language model agents have evolved from passive text generators into active, tool-using systems. ReAct~\cite{yao2022react} established the paradigm of interleaving reasoning with tool interactions, enabling agents to plan, act, and adapt. Reflexion~\cite{shinn2023reflexion} and Self-Refine~\cite{madaan2023self} introduced feedback loops, treating tool calls as recoverable actions subject to iterative correction. Subsequent work expanded tool coverage. Toolformer~\cite{schick2023toolformer} trained models to invoke APIs, ToolLLM~\cite{qin2024toolllm} generalized across 16,000 real-world APIs, and HuggingGPT~\cite{shen2023hugginggpt} showed that language can orchestrate AI models. More recently, Search-R1~\cite{jin2025search} trained agents via reinforcement learning to interleave reasoning with real-time retrieval, A-Mem~\cite{xu2026mem} introduced dynamic agentic memory for long-horizon tasks, and MCP-Zero~\cite{fei2025mcp, hou2025model, ray2025survey} enabled on-demand tool discovery at scale. Frameworks such as OpenClaw~\cite{openclaw2025} integrate these advances into a modular runtime~\cite{wang2024survey, xi2025rise}. Despite their versatility, these frameworks share an implicit assumption that execution is recoverable. This assumption holds in conventional settings such as information retrieval and software automation~\cite{jimenez2024swe, yang2024swe, zhou2024webarena}, but breaks down in blockchain environments.

\textit{On-Chain Transaction Safety.}
A growing body of work prevents unsafe execution beforehand, intervening on either the code or the agent that issues it. The first secures the on-chain artifact. At the contract level, GPTScan~\cite{sun2024gptscan} surfaces logic flaws, SMARTINV~\cite{wang2024smartinv} infers cross-modal invariants, and CrossGuard~\cite{zhang2025crossguard} enforces control-flow integrity at runtime. At the transaction level, Theorem-Carrying Transactions~\cite{bjorner2024theorem} attach proofs verified before execution, and MEV research~\cite{zhou2023sok, torres2024rolling} counters front-running and sandwiching via private channels like Flashbots Protect~\cite{alqithami2026autonomous}. The second guards the agent: GuardAgent~\cite{xiang2024guardagent} compiles safety requests into guardrail code, ShieldAgent~\cite{chen2025shieldagent} verifies action trajectories against policy-derived rule circuits, AGrail~\cite{luo2025agrail} adaptively synthesizes checks that block risky actions, and AgentSpec~\cite{wang2025agentspec} enforces user-defined rules at runtime. Yet these guardrails are domain-agnostic, assuming a recoverability that on-chain irreversibility denies: the gap ChainClaw's pre-execution safety pipeline closes.

\begin{figure*}[t]
\centering
\includegraphics[scale=0.32]{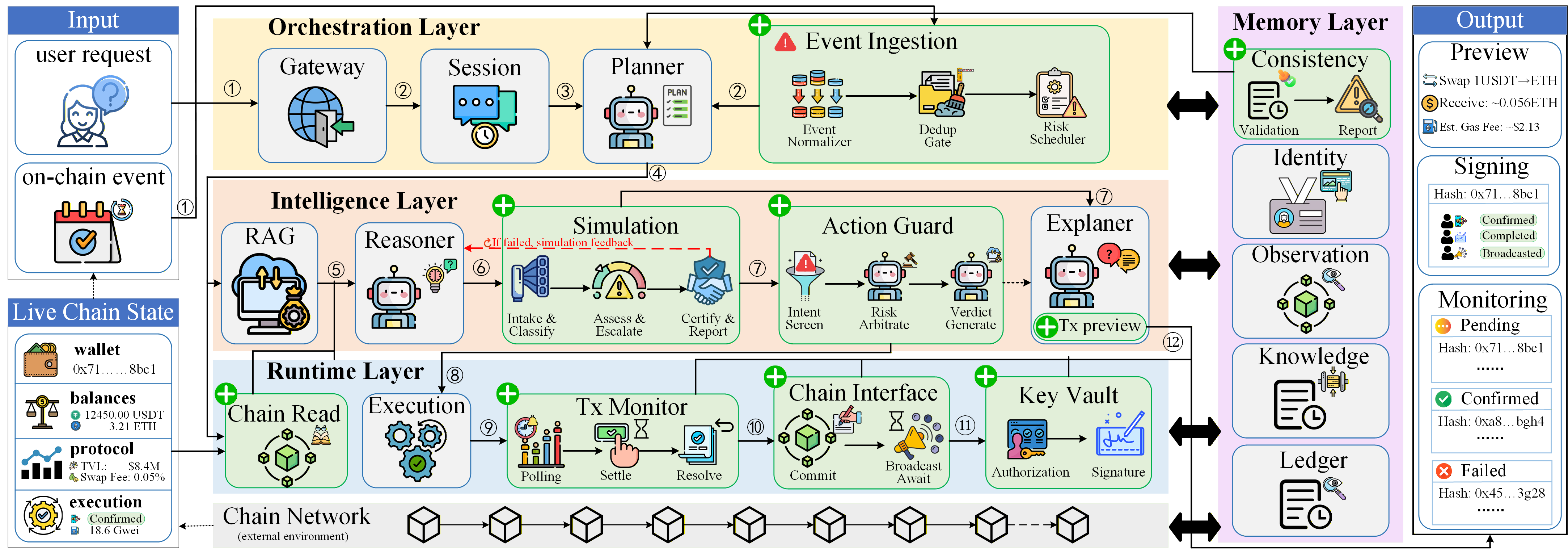}
\caption{Overall architecture of ChainClaw, a layered pre-execution safety pipeline for on-chain LLM agents. Gray modules are inherited from OpenClaw, those marked with a green plus are newly introduced in this work. All modules are reorganized into five layers: orchestration, intelligence, runtime, memory and chain network.}
\label{method}
\vspace{-1em}
\end{figure*}

\textit{Blockchain-Native Agents.}
Research linking LLM agents and blockchain remains nascent and fragmented; we group it into three strands. \textit{Chain access as tools.} LangChain-style toolkits~\cite{langchain2022} expose chain data and contract calls to agents, and Web3Agent~\cite{fan2025web3agent} decomposes user intents into on-chain workflows. \textit{Domain-specific autonomy.} For trading, CryptoTrade~\cite{li2024cryptotrade} fuses on- and off-chain signals through a reflective loop, and Agent Market Arena~\cite{qian2025agents} benchmarks agents under live market conditions; for governance, Auto.gov~\cite{xu2025auto} tunes DeFi parameters on-chain. \textit{On-chain multi-agent coordination.} Smart contracts anchor decentralized collaboration~\cite{jin2024decoagent}, harden multi-agent consensus against Byzantine peers~\cite{chen2024blockagents}, and incentivize agent coordination~\cite{qi2026towards}. Across all three strands, safety stays bolted on per task; ChainClaw instead bakes it into a unified, layered architecture for the full on-chain execution lifecycle.

\section{Method}

\textit{Overview.} As shown in Figure~\ref{method}, ChainClaw builds on OpenClaw and reorganizes components into five cooperating layers (orchestration, intelligence, runtime, a cross-cutting memory layer, and chain network), so every action is grounded in live chain state and screened before it touches an irreversible ledger. The orchestration layer accepts two kinds of input: a user request flows through the Gateway, Session, and Planner pipeline, whereas on-chain events are read from live chain state via Chain Read and routed through Event Ingestion into the same Planner. Drawing on RAG-retrieved knowledge and the chain state from Chain Read, the Planner passes the task to the Reasoner. The resulting action first undergoes Simulation, ChainClaw's core safeguard, which tests it before execution and returns control to the Reasoner on failure, then proceeds to the Action Guard and Explainer; if not blocked, it advances to Execution, tracked by the Tx Monitor, and is written through the unified Chain Interface with signing protected by the Key Vault. These stages map to the system outputs: the Explainer yields the preview, the Chain Interface and Key Vault the signing, and the Tx Monitor the monitoring. Unlike OpenClaw, ChainClaw adds a Consistency module that aligns the Planner's output with on-chain state, while the memory layer persists identity, observation, knowledge, and ledger records throughout.

\vspace{-1em}
\subsection{Orchestration layer for Reactivity}
The orchestration layer formalizes the system's input space and closes the Reactivity gap with a dual-trigger execution model over heterogeneous inputs. Blockchain is not static but a continuously evolving system whose state changes are externally driven, frequent, and often irreversible; an agent must therefore participate in on-chain dynamics rather than react to user commands alone.

\textit{Dual-trigger input.} ChainClaw operates over two input classes: (i) user-driven requests, synchronous and issued through the Gateway, and (ii) chain-driven signals, derived from the evolving blockchain state rather than directly observed events. For the latter, ChainClaw maintains a continuously updated \textit{live chain state}, grounded in the Chain Network and accessed through Chain Read, giving a structured interpretation of the execution environment rather than treating raw logs as inputs. The live chain state has four complementary categories:
\textbf{Network-level state}: the underlying Chain Network, providing the global execution context for all state transitions.
\textbf{Wallet-level state}: account-level conditions such as balances and available assets.
\textbf{Protocol-level state}: smart contract status and authorization relations governing executable actions.
\textbf{Execution-level state}: transaction lifecycle dynamics (pending, confirmed, failed).
From temporal changes in this state, the Event Ingestion pipeline derives chain-driven signals, converting state transitions into normalized events for downstream reasoning.

\textit{Dual-path input processing.} ChainClaw instantiates two asymmetric paths by input origin, merged into a unified Planner interface so a single reasoning core spans synchronous and asynchronous inputs. \textit{User-driven requests} follow the OpenClaw route: the \textbf{Gateway} handles authentication and routing, and a \textbf{Session} module maintains short-term context before the Planner, preserving compatibility with LLM-agent workflows. \textit{Chain-driven inputs} instead pass through an \textbf{Event Ingestion} pipeline for asynchronous environments, handling on-chain signals such as exploit attempts or transaction failures while mitigating overload. It comprises the Event Normalizer (heterogeneous logs into a unified representation), the Dedup Gate (removing duplicates from multi-node propagation), and the Risk Scheduler (priority by urgency and impact), plus a threshold-based circuit breaker that buffers excess events and reintroduces them once load stabilizes.

\textit{Planning and state-consistent.} The \textbf{Planner} decomposes each task into an executable plan. Because chain state evolves continuously, a plan can drift between when it is composed and when its steps take effect; and because a task spans modules (guard, execution, monitoring), their records can conflict. A purely structural check catches neither failure, so ChainClaw pairs the Planner with a \textbf{Consistency} module that runs over the task's accumulated memory rather than a one-off validation pass. It reads the on-chain state and retrieves the task's execution trace (transaction previews, guard decisions, execution outcomes, monitoring results), then validates it structurally (schema, step ordering, field completeness, task status) and semantically (cross-module conflicts), issuing one verdict that lets it proceed or triggers replan, retry, or termination.

\subsection{Intelligence layer for Irreversibility}
\textit{Reasoning and simulation.} Given a task plan from the Planner, ChainClaw gathers context through two parallel channels: the \textbf{RAG module} for external blockchain knowledge and \textbf{Chain Read} for live on-chain state, together forming a grounded view of the execution environment. Conditioned on this context, the Reasoner deliberates over the plan, so reasoning is grounded in both prior knowledge and the latest chain state rather than isolated. The Reasoner is restricted to inference only, with no access to private keys or signing, and so cannot trigger state-changing actions directly. Its candidate actions then enter the Simulation stage, ChainClaw's core pre-execution safeguard: because blockchain execution is irreversible, simulation is mandatory before any commitment, and an unsafe outcome returns control to the Reasoner for replanning.

The \textbf{Simulation} module operates in three stages: \textit{Intake and classification} aggregates the relevant context (chain state, prior analysis, and the candidate operation) and labels the action as passive (observation-level) or state-changing (execution-level), setting the sensitivity of downstream checks.
\textit{Risk assessment and escalation} scores the action against behavioral heuristics (anomalous transaction patterns, cross-chain activity, abnormal frequency), yielding a conservative score that can only increase under further signals.
\textit{Certification and reporting} verifies data completeness and emits a structured report that flags incomplete chain data and summarizes risk levels, triggered rules, and side effects, then forwards it to the Action Guard for authorization.

\textit{Action Guard.} Actions that pass Simulation reach the \textbf{Action Guard}, which filters residual unsafe behaviors before any state-changing operation. It has three components:
\textit{Intent screen} applies hard rule-based filters on the simulation output, removing malicious or disallowed actions via deterministic constraints.
\textit{Risk arbitrate} runs an LLM-based second pass over the coarse risk label (high, medium, low) from upstream rules; since such labels lose nuance, it reweighs the address's historical behavior, market conditions, and consistency between intent and chain state to catch ambiguous cases hard rules miss.
\textit{Verdict generation} emits an interpretable decision (why an action is blocked, which factors triggered it, and possible next steps), improving transparency for users and auditors without weakening safety guarantees.

\begin{table*}[t]
\centering
\caption{Overall performance comparison across seven blockchain-agent tasks.}
\label{tab:main-results}
\setlength{\tabcolsep}{3pt}
\renewcommand{\arraystretch}{1.15}
\footnotesize
\begin{tabular}{l*{25}{c}}
\toprule
& \multicolumn{2}{c}{T1} & \multicolumn{2}{c}{T2}
& \multicolumn{3}{c}{T3} & \multicolumn{3}{c}{T4}
& \multicolumn{5}{c}{T5} & \multicolumn{5}{c}{T6}
& \multicolumn{5}{c}{T7} \\
\cmidrule(lr){2-3}\cmidrule(lr){4-5}\cmidrule(lr){6-8}\cmidrule(lr){9-11}%
\cmidrule(lr){12-16}\cmidrule(lr){17-21}\cmidrule(lr){22-26}
Method & \#CP & \#CR & \#CP & \#CR & \#CP & \#CR & \#SF & \#CP & \#CR & \#SF
& \#CP & \#CR & \#SF & \#RB & \#EF & \#CP & \#CR & \#SF & \#RB & \#EF & \#CP & \#CR & \#SF & \#RB & \#EF \\
\midrule
ReAct      & 1.00 & 0.67 & 1.00 & 0.67 & 0.29 & 0.25 & 0.00 & 0.29 & 0.50 & 0.00 & 0.33 & 0.29 & 0.00 & 1.00 & 0.29 & 0.50 & 0.33 & 0.33 & 1.00 & 0.50 & 0.67 & 0.67 & 1.00 & 0.67 & 0.67 \\
LangChain  & 1.00 & 0.67 & 1.00 & 0.67 & 0.43 & 0.75 & 0.50 & 0.57 & 0.75 & 0.60 & 0.67 & 0.86 & 0.40 & 0.67 & 0.67 & 0.50 & 0.33 & 0.50 & 1.00 & 0.50 & 0.67 & 0.67 & 1.00 & 0.67 & 0.67 \\
OpenClaw   & 1.00 & 0.67 & 1.00 & 0.67 & 0.86 & 0.25 & 0.75 & 0.86 & 0.50 & 0.80 & 0.89 & 0.71 & 0.80 & 0.67 & 0.89 & 0.50 & 0.67 & 0.00 & 1.00 & 0.36 & 0.67 & 0.67 & 1.00 & 0.67 & 0.67 \\
\textbf{ChainClaw} & \textbf{1.00} & \textbf{1.00} & \textbf{1.00} & \textbf{1.00} & \textbf{1.00} & \textbf{1.00} & \textbf{1.00} & \textbf{1.00} & \textbf{1.00} & \textbf{1.00} & \textbf{1.00} & \textbf{1.00} & \textbf{1.00} & \textbf{1.00} & \textbf{1.00} & \textbf{1.00} & \textbf{1.00} & \textbf{1.00} & \textbf{1.00} & \textbf{1.00} & \textbf{1.00} & \textbf{1.00} & \textbf{1.00} & \textbf{1.00} & \textbf{1.00} \\
\bottomrule
\end{tabular}
\end{table*}

\begin{figure*}[t]
\centering
\includegraphics[scale=0.32]{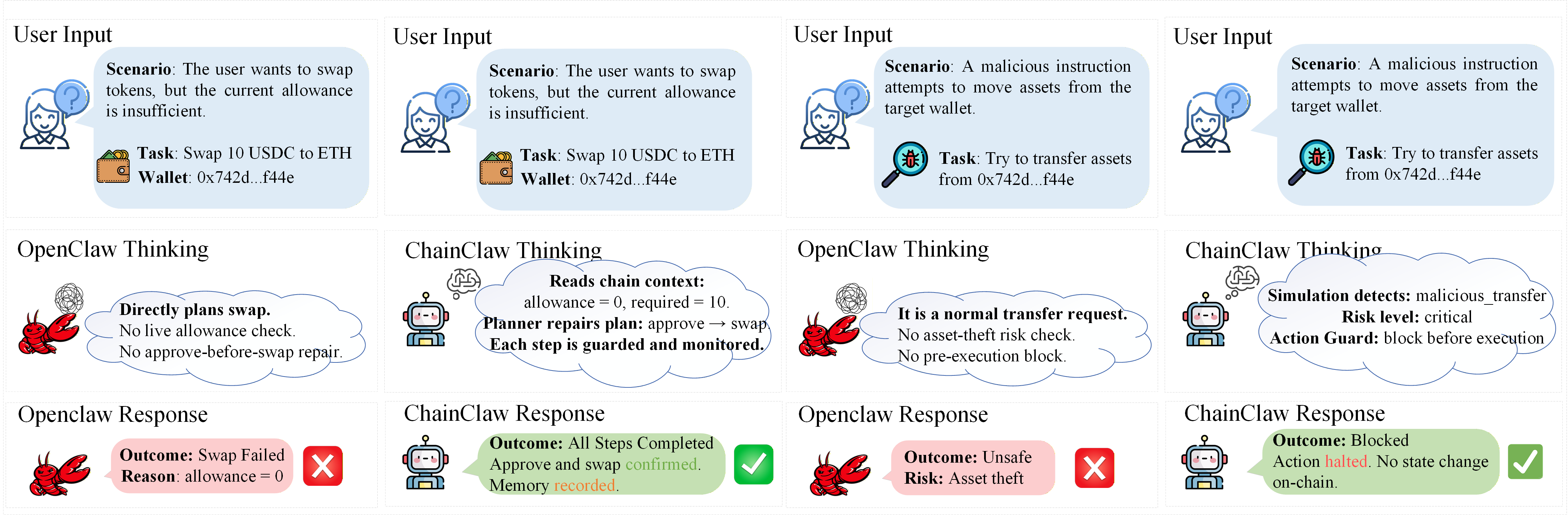}
\caption{Case study of ChainClaw vs.OpenClaw on a complex swap (left) and a malicious transfer (right). OpenClaw plans directly and fails or executes unsafely; ChainClaw repairs with chain context and blocks attacks pre-execution.}
\label{case}
\vspace{-1em}
\end{figure*}

\textit{Explainer and transaction preview.} The \textbf{Explainer}, the pipeline's final user-facing stage, consolidates upstream information (the Reasoner's intent, the simulation outcome, and above all the Action Guard's verdict) into a human-readable report issued before anything is committed on-chain. Its form follows the verdict, taking one of two modes:
\textit{Blocked.} When the Action Guard blocks an action, the Explainer reports the decision and rationale, stating which check failed (for instance, unsafe asset movement or abnormal approval), so the user understands why execution was halted.
\textit{Cleared.} When the action is cleared, it renders a transaction preview that describes, in natural language, what the operation will do, its expected output, and its estimated gas cost; purely informational requests that change no on-chain state terminate here, answered directly without signing.
Because this preview appears before any signature is requested, the user sees exactly what a transaction does and costs before authorizing it, rather than committing to an opaque payload.

\vspace{-1em}
\subsection{Runtime layer for Observability}
\textit{Transaction monitor.} The \textbf{Execution} module broadcasts approved commands from the Action Guard to the blockchain. Because execution is asynchronous with variable confirmation latency (e.g., Bitcoin's 10-minute blocks), ChainClaw adds a \textbf{Transaction Monitor} that tracks the transaction lifecycle in real time over three stages. \textit{Polling} starts tracking by recording the submission timestamp, setting a timeout, and periodically querying the node for status. \textit{Settlement tracking} follows progression, waiting for receipt inclusion (successful mining) and accumulating confirmations until a safety threshold is reached. \textit{Resolution} inspects the execution status and labels the transaction \textit{confirmed} on sufficient confirmations or \textit{failed} on revert or timeout, packaging the result for upstream use. The monitor also streams intermediate updates to the interface, exposing transaction hashes and statuses (pending, confirmed, failed), giving real-time visibility for auditing on-chain behavior.

\textit{Chain interface and key vault.} Once approved upstream, ChainClaw executes state updates through a unified \textbf{Chain Interface} that abstracts heterogeneity across blockchains and gives a consistent write layer for all backends, in two stages: \textit{commit} applies the state transition by permanently recording the transaction on-chain, and \textit{broadcast and await} submits it to the network and tracks propagation until execution feedback is observed, so downstream components see both submission and inclusion uniformly. For secure authorization, ChainClaw adds a \textbf{Key Vault} that isolates private-key management from reasoning and execution, with two functions: \textit{authorization} checks whether the requesting component may initiate a signature, enforcing strict access control, and \textit{signature} signs cryptographically while keeping keys encapsulated, never exposing key material to external modules. Acting together, the Key Vault signs and the Chain Interface broadcasts, surfacing a concise confirmation: \emph{Outcome: Signed and broadcast. Transaction submitted to Ethereum mainnet.}

\textit{Memory.} ChainClaw's memory largely follows OpenClaw, except for the new Consistency module tightly integrated with the orchestration layer. Beyond this, the memory is redesigned as a shared, cross-layer persistent store accessible to the orchestration, intelligence, and runtime layers for bidirectional retrieval and state reconstruction. It supports four core capabilities: (i) persistence, storing execution records durably for recovery after restarts; (ii) cross-session memory, preserving interaction context across sessions; (iii) semantic retrieval, enabling similarity-based access to past experiences beyond key-based lookup; and (iv) compression and decay, periodically summarizing older records to bound growth while retaining essential information.

\section{Experiments}

\begin{table*}[t]
\centering
\caption{Ablation study grouped by the three motivations: Rea. (Reactivity), Irr. (Irreversibility), and Obs. (Observability).}
\label{tab:ablation}
\setlength{\tabcolsep}{2.5pt}
\renewcommand{\arraystretch}{1.1}
\footnotesize
\begin{tabular}{@{}c@{\hspace{2pt}}l*{25}{c}@{}}
\toprule
& & \multicolumn{2}{c}{T1} & \multicolumn{2}{c}{T2}
& \multicolumn{3}{c}{T3} & \multicolumn{3}{c}{T4}
& \multicolumn{5}{c}{T5} & \multicolumn{5}{c}{T6} & \multicolumn{5}{c}{T7} \\
\cmidrule(lr){3-4}\cmidrule(lr){5-6}\cmidrule(lr){7-9}\cmidrule(lr){10-12}%
\cmidrule(lr){13-17}\cmidrule(lr){18-22}\cmidrule(lr){23-27}
& Method & \#CP & \#CR & \#CP & \#CR & \#CP & \#CR & \#SF & \#CP & \#CR & \#SF
& \#CP & \#CR & \#SF & \#RB & \#EF & \#CP & \#CR & \#SF & \#RB & \#EF & \#CP & \#CR & \#SF & \#RB & \#EF \\
\midrule
\multirow{4}{*}{\rotatebox[origin=c]{90}{\textbf{Rea.}}}
& w/o Event Ingestion        & 1.00 & 1.00 & 1.00 & 1.00 & 1.00 & 1.00 & 1.00 & 1.00 & 1.00 & 1.00 & 1.00 & 1.00 & 1.00 & 1.00 & 1.00 & 1.00 & 1.00 & 1.00 & 1.00 & 1.00 & 0.17 & 0.00 & 0.20 & 0.20 & 0.50 \\
& Session-only handling & 1.00 & 1.00 & 1.00 & 1.00 & 1.00 & 1.00 & 1.00 & 1.00 & 1.00 & 1.00 & 1.00 & 1.00 & 1.00 & 1.00 & 1.00 & 1.00 & 1.00 & 1.00 & 1.00 & 1.00 & 0.70 & 0.58 & 0.75 & 0.55 & 0.70 \\
& w/o Sim. Feedback & 1.00 & 1.00 & 1.00 & 1.00 & 1.00 & 1.00 & 1.00 & 1.00 & 1.00 & 1.00 & 0.92 & 0.88 & 1.00 & 0.55 & 0.82 & 1.00 & 0.92 & 1.00 & 0.58 & 0.78 & 1.00 & 1.00 & 1.00 & 1.00 & 1.00 \\
& One-shot reasoning         & 1.00 & 1.00 & 1.00 & 1.00 & 1.00 & 1.00 & 1.00 & 1.00 & 1.00 & 1.00 & 0.90 & 0.84 & 0.90 & 0.62 & 0.78 & 0.95 & 0.86 & 0.92 & 0.62 & 0.74 & 0.92 & 0.88 & 0.92 & 0.72 & 0.82 \\
\midrule
\multirow{4}{*}{\rotatebox[origin=c]{90}{\textbf{ Irr.}}}
& w/o Simulation             & 1.00 & 1.00 & 1.00 & 1.00 & 1.00 & 0.72 & 0.70 & 1.00 & 0.72 & 0.70 & 0.92 & 0.72 & 0.70 & 0.55 & 0.82 & 1.00 & 0.65 & 0.65 & 0.55 & 0.75 & 1.00 & 0.67 & 0.60 & 0.60 & 1.00 \\
& Simple simulation          & 1.00 & 1.00 & 1.00 & 1.00 & 1.00 & 0.85 & 0.82 & 1.00 & 0.88 & 0.85 & 1.00 & 0.85 & 0.82 & 0.78 & 1.00 & 1.00 & 0.82 & 0.80 & 0.70 & 0.92 & 1.00 & 0.83 & 0.80 & 0.60 & 1.00 \\
& w/o Action Guard           & 1.00 & 1.00 & 1.00 & 1.00 & 1.00 & 1.00 & 0.40 & 1.00 & 1.00 & 0.45 & 1.00 & 1.00 & 0.40 & 1.00 & 1.00 & 1.00 & 0.70 & 0.00 & 0.35 & 0.40 & 1.00 & 1.00 & 0.70 & 1.00 & 1.00 \\
& Rule-only Guard            & 1.00 & 1.00 & 1.00 & 1.00 & 1.00 & 0.90 & 0.70 & 1.00 & 0.92 & 0.75 & 1.00 & 0.90 & 0.70 & 0.82 & 1.00 & 1.00 & 0.82 & 0.60 & 0.60 & 0.80 & 1.00 & 1.00 & 0.85 & 1.00 & 1.00 \\
\midrule
\multirow{5}{*}{\rotatebox[origin=c]{90}{\textbf{Obs. }}}
& w/o On-chain Read          & 0.50 & 0.33 & 0.80 & 0.75 & 0.92 & 0.76 & 0.85 & 0.92 & 0.78 & 0.85 & 0.92 & 0.76 & 0.85 & 0.67 & 0.90 & 0.90 & 0.72 & 0.75 & 0.62 & 0.85 & 1.00 & 1.00 & 1.00 & 1.00 & 1.00 \\
& Stable On-chain Read       & 1.00 & 0.78 & 1.00 & 0.72 & 0.96 & 0.82 & 0.90 & 0.96 & 0.84 & 0.90 & 0.96 & 0.78 & 0.90 & 0.72 & 0.92 & 0.96 & 0.80 & 0.82 & 0.70 & 0.88 & 1.00 & 1.00 & 1.00 & 1.00 & 1.00 \\
& Partial On-chain Read      & 0.88 & 0.82 & 0.60 & 0.50 & 0.94 & 0.78 & 0.88 & 0.94 & 0.80 & 0.88 & 0.94 & 0.78 & 0.88 & 0.68 & 0.92 & 0.94 & 0.78 & 0.80 & 0.66 & 0.86 & 1.00 & 1.00 & 1.00 & 1.00 & 1.00 \\
& w/o Tx Monitor             & 1.00 & 1.00 & 1.00 & 1.00 & 0.88 & 1.00 & 1.00 & 0.90 & 1.00 & 1.00 & 0.88 & 1.00 & 1.00 & 0.45 & 1.00 & 1.00 & 1.00 & 1.00 & 1.00 & 1.00 & 1.00 & 1.00 & 1.00 & 0.55 & 1.00 \\
& Receipt-only Monitor       & 1.00 & 1.00 & 1.00 & 1.00 & 1.00 & 1.00 & 1.00 & 1.00 & 1.00 & 1.00 & 1.00 & 1.00 & 1.00 & 0.50 & 1.00 & 1.00 & 1.00 & 1.00 & 1.00 & 1.00 & 0.95 & 1.00 & 1.00 & 0.65 & 0.95 \\
\midrule
& \textbf{Full ChainClaw} & \textbf{1.00} & \textbf{1.00} & \textbf{1.00} & \textbf{1.00} & \textbf{1.00} & \textbf{1.00} & \textbf{1.00} & \textbf{1.00} & \textbf{1.00} & \textbf{1.00} & \textbf{1.00} & \textbf{1.00} & \textbf{1.00} & \textbf{1.00} & \textbf{1.00} & \textbf{1.00} & \textbf{1.00} & \textbf{1.00} & \textbf{1.00} & \textbf{1.00} & \textbf{1.00} & \textbf{1.00} & \textbf{1.00} & \textbf{1.00} & \textbf{1.00} \\
\bottomrule
\end{tabular}
\end{table*}

\subsection{Experimental Settings}
\textit{Benchmark Tasks.}
Our benchmark comprises seven tasks (T1--T7) organized into four categories. \textit{Read} tasks query live chain state without modifying it: retrieving the ETH balance of an address (T1) and the ERC-20 allowance granted to a spender (T2). \textit{Single-step execution} tasks construct one state-changing transaction: transferring 10~USDC to a specified address (T3) and approving 100~USDC to a contract (T4). \textit{Multi-step execution} requires composing dependent actions: when allowance is insufficient, the agent must first approve and then swap (T5). \textit{Security/anomaly} tasks probe the safety boundary: an adversarial instruction that attempts to drain assets, which a correct agent must refuse (T6), and a passively triggered on-chain event to which the agent must react appropriately (T7).

\textit{Evaluation Metrics.}
We assess performance with five metrics, each normalized to [0,1]. \textit{Completion} (\textbf{\#CP}), the fraction of required steps an agent carries through, is computed as $\#\mathrm{CP}$=(completed steps)/(total steps), averaged over instances and applicable to all tasks. \textit{Correctness} (\textbf{\#CR}), applicable to all tasks, measures how closely the produced output or constructed operation matches the expected result: field-level F1 for read tasks, and transaction-parameter or operation correctness for execution tasks. \textit{Safety} (\textbf{\#SF}), the agent's ability to block illegal operations, is computed as $\#\mathrm{SF}$=1-(unblocked illegal operations)/(potential illegal operations), defined over the execution and anomaly tasks (T3--T7). \textit{Robustness} (\textbf{\#RB}), stable execution under anomalous or asynchronous conditions, is computed as $\#\mathrm{RB}$=1-(failed executions)/(total anomalous events), defined over the multi-step and passively triggered tasks (T5, T7). Finally, \textit{Efficiency} (\textbf{\#EF}) compares actual usage against a per-task threshold, computed as $\#\mathrm{EF}$=1-(actual resource or time)/(threshold resource or time), where usage covers gas, step count, and number of transactions; it applies to T3--T5 and T7. Each reported score is averaged over the $N$ instances within its task category.

\textit{Baselines.}
We compare ChainClaw against four representative agent frameworks that span the three tiers of our taxonomy: \textit{(1) orchestration frameworks.} OpenClaw~\cite{openclaw2025}, the general-purpose framework that ChainClaw builds upon, and LangChain~\cite{langchain}; \textit{(2) autonomous reasoning agents.} ReAct~\cite{yao2022react}. For fairness, \textit{Completion} excludes steps a baseline cannot support, while \textit{Safety} and \textit{Robustness} are scored over all cases.

\textit{Execution Environment and Implementation Details.}
Read tasks (T1--T2) are executed against live Ethereum mainnet through an Alchemy JSON-RPC endpoint, with contract metadata resolved via the Etherscan API; the balances and allowances they return reflect real on-chain state at query time. For state-changing tasks (T3--T7), the pipeline runs through to transaction preview without broadcasting, so gas is estimated rather than realized and \textit{Safety} is judged at the pre-signing stage. All systems share a single backbone, \texttt{moonshot-v1-8k}, queried with temperature 0.3 and a maximum of 1024 output tokens; these decoding settings are held fixed across ChainClaw and every baseline, and inference is performed through the provider's API without local GPU training. The RAG module is backed by a small corpus of blockchain-related knowledge collected from public web sources.

\subsection{Main results}

\begin{table}[t]
\centering
\caption{User study results (mean $\pm$ std; higher is better).}
\label{tab:user-study}
\small
\setlength{\tabcolsep}{4pt}
\resizebox{\columnwidth}{!}{%
\begin{tabular}{lccccc}
\toprule
System & Usefulness & Safety & Trust & Controllability & Preference \\
\midrule
OpenClaw & $4.72 \pm 0.91$ & $4.18 \pm 1.06$ & $4.35 \pm 0.98$ & $4.51 \pm 0.94$ & 27.8\% \\
\textbf{ChainClaw} & $\mathbf{6.08} \pm 0.63$ & $\mathbf{6.31} \pm 0.55$ & $\mathbf{5.96} \pm 0.68$ & $\mathbf{6.17} \pm 0.61$ & \textbf{72.2\%} \\
\bottomrule
\end{tabular}}
\vspace{-2em}
\end{table}

\textit{Overall Performance.} Table~\ref{tab:main-results} reports all four systems on seven tasks. ChainClaw achieves 1.00 on every metric, while the baselines match it only on read queries and degrade on state-changing and event-driven tasks; since all share the \texttt{moonshot-v1-8k} backbone and decoding settings, the differences reflect architecture, not model capability. On reads (T1--T2), all systems complete the queries (\#CP=1.00) but the baselines stay at \#CR=0.67 against ChainClaw's 1.00. On transaction construction (T3--T4), ReAct collapses (\#CP=0.29, \#SF=0.00), and OpenClaw, despite \#CP=0.86, drops to \#CR=0.25 on T3 with only partial safety. On the adversarial drain (T6), OpenClaw fails to block it (\#SF=0.00) while ChainClaw refuses it (\#SF=1.00); on the passive event (T7), all baselines plateau at 0.67/0.67/1.00/0.67/0.67, and ChainClaw reaches 1.00 on both. These failures match our three motivations, which ChainClaw closes by architecture rather than a stronger model.

\textit{Case study.} Qualitative examples in Figure~\ref{fig:case-study} show ChainClaw's practical advantages. For insufficient allowance, ChainClaw reads live state, repairs the plan into approve-then-swap, and completes execution, while OpenClaw fails. For malicious transfer, ChainClaw simulates and blocks the action before signing, while OpenClaw proceeds unsafely. These cases demonstrate ChainClaw's gains in Observability, Reactivity, and Irreversibility.

\textit{User study.} We complement the automatic evaluation with a user study: 12 participants rated an OpenClaw-style baseline and ChainClaw across 3 scenarios on five seven-point Likert dimensions and a forced preference choice (Table~\ref{tab:user-study}). ChainClaw wins on every dimension, with the largest gains on Safety (6.31 vs.\ 4.18) and Controllability (6.17 vs.\ 4.51), the dimensions most tied to pre-execution risk, and clear margins on Trust (5.96 vs.\ 4.35) and Usefulness (6.08 vs.\ 4.72). Its lower variance (std 0.55--0.68 vs.\ 0.91--1.06) indicates broad agreement, and 72.2\% prefer it overall. User perceptions mirror the metrics: ChainClaw's gains are not only measured but felt.

\subsection{Ablation study}

We ablate the modules behind each motivation by removing or weakening one component at a time and compare them against full ChainClaw, which scores 1.00 throughout (Table~\ref{tab:ablation}).

\textit{Reactivity.} Reactive capability depends on event ingestion and simulation feedback. Removing event ingestion causes the passively triggered task to collapse (T7 \#CP=0.17, \#CR=0.00), while session-only handling recovers only partially (T7 \#CP=0.70, \#RB=0.55). Without simulation feedback, robustness drops on multi-step and drain tasks (\#RB=0.55 on T5 and 0.58 on T6), and one-shot reasoning similarly degrades T5 (0.90/0.84/0.90/0.62/0.78). Thus, Reactivity requires both an asynchronous event path for T7 and an iterative simulate-then-replan loop for T5--T6.

\textit{Irreversibility.} The pre-execution safety pipeline relies on both simulation and the action guard. Removing simulation lowers correctness and safety on state-changing tasks (T3 \#CR=0.72, \#SF=0.70; T6 \#SF=0.65), and simplified simulation only partially recovers safety (T6 \#SF=0.80). Removing the action guard preserves completion but severely reduces safety, especially on the malicious drain where T6 \#SF falls to 0.00. A rules-only guard also remains insufficient (T6 \#SF=0.60). These results show that effective blocking requires simulation evidence and guard-level enforcement together.

\textit{Observability.} Observability depends on live on-chain reads and transaction monitoring. Removing on-chain read hurts both read tasks (T1 \#CP=0.50, \#CR=0.33) and state-dependent execution (T6 \#CR=0.72); weaker reads still cap correctness, e.g., stable read gives T1 \#CR=0.78 and partial read gives T2 \#CP=0.60. Removing the transaction monitor mainly affects robustness, reducing \#RB to 0.45 on T5 and 0.55 on T7, while receipt-only monitoring remains low (T5 \#RB=0.50). Full observability therefore requires live reads for correctness and continuous monitoring for robust recovery.

\section{Conclusion}
We presented ChainClaw, an on-chain agent framework that closes the Reactivity, Irreversibility, and Observability gaps left by general-purpose runtimes, through an event-driven orchestration layer, a simulation-based safety intelligence layer, and an on-chain monitoring runtime layer, unified by a cross-layer memory. The results indicate that reliable blockchain-native agency requires a dedicated architecture rather than a stronger model.

\bibliographystyle{ACM-Reference-Format}
\bibliography{sample-base}


\begin{thebibliography}{43}


\ifx \showCODEN    \undefined \def \showCODEN     #1{\unskip}     \fi
\ifx \showISBNx    \undefined \def \showISBNx     #1{\unskip}     \fi
\ifx \showISBNxiii \undefined \def \showISBNxiii  #1{\unskip}     \fi
\ifx \showISSN     \undefined \def \showISSN      #1{\unskip}     \fi
\ifx \showLCCN     \undefined \def \showLCCN      #1{\unskip}     \fi
\ifx \shownote     \undefined \def \shownote      #1{#1}          \fi
\ifx \showarticletitle \undefined \def \showarticletitle #1{#1}   \fi
\ifx \showURL      \undefined \def \showURL       {\relax}        \fi
\providecommand\bibfield[2]{#2}
\providecommand\bibinfo[2]{#2}
\providecommand\natexlab[1]{#1}
\providecommand\showeprint[2][]{arXiv:#2}

\bibitem[Alqithami(2026)]%
        {alqithami2026autonomous}
\bibfield{author}{\bibinfo{person}{Saad Alqithami}.} \bibinfo{year}{2026}\natexlab{}.
\newblock \showarticletitle{Autonomous Agents on Blockchains: Standards, Execution Models, and Trust Boundaries}.
\newblock \bibinfo{journal}{\emph{arXiv preprint arXiv:2601.04583}} (\bibinfo{year}{2026}).
\newblock


\bibitem[Babel et~al\mbox{.}(2023)]%
        {babel2023clockwork}
\bibfield{author}{\bibinfo{person}{Kushal Babel}, \bibinfo{person}{Philip Daian}, \bibinfo{person}{Mahimna Kelkar}, {and} \bibinfo{person}{Ari Juels}.} \bibinfo{year}{2023}\natexlab{}.
\newblock \showarticletitle{Clockwork finance: Automated analysis of economic security in smart contracts}. In \bibinfo{booktitle}{\emph{2023 IEEE Symposium on Security and Privacy (SP)}}. IEEE, \bibinfo{pages}{2499--2516}.
\newblock


\bibitem[Bj{\o}rner et~al\mbox{.}(2024)]%
        {bjorner2024theorem}
\bibfield{author}{\bibinfo{person}{Nikolaj~S Bj{\o}rner}, \bibinfo{person}{Ashley~J Chen}, \bibinfo{person}{Shuo Chen}, \bibinfo{person}{Yang Chen}, \bibinfo{person}{Zhongxin Guo}, \bibinfo{person}{Tzu-Han Hsu}, \bibinfo{person}{Peng Liu}, {and} \bibinfo{person}{Nanqing Luo}.} \bibinfo{year}{2024}\natexlab{}.
\newblock \showarticletitle{Theorem-Carrying-Transaction: Runtime Certification to Ensure Safety for Smart Contract Transactions.}
\newblock \bibinfo{journal}{\emph{CoRR}} (\bibinfo{year}{2024}).
\newblock


\bibitem[Chase(2022a)]%
        {langchain2022}
\bibfield{author}{\bibinfo{person}{Harrison Chase}.} \bibinfo{year}{2022}\natexlab{a}.
\newblock \bibinfo{title}{{LangChain}}.
\newblock \bibinfo{howpublished}{\url{https://github.com/langchain-ai/langchain}}.
\newblock
\newblock
\shownote{Open-source framework for LLM application development; accessed 2026-06-13}.


\bibitem[Chase(2022b)]%
        {langchain}
\bibfield{author}{\bibinfo{person}{Harrison Chase}.} \bibinfo{year}{2022}\natexlab{b}.
\newblock \bibinfo{title}{{LangChain}}.
\newblock \bibinfo{howpublished}{\url{https://github.com/langchain-ai/langchain}}.
\newblock


\bibitem[Chen et~al\mbox{.}(2024)]%
        {chen2024blockagents}
\bibfield{author}{\bibinfo{person}{Bei Chen}, \bibinfo{person}{Gaolei Li}, \bibinfo{person}{Xi Lin}, \bibinfo{person}{Zheng Wang}, {and} \bibinfo{person}{Jianhua Li}.} \bibinfo{year}{2024}\natexlab{}.
\newblock \showarticletitle{Blockagents: Towards byzantine-robust llm-based multi-agent coordination via blockchain}. In \bibinfo{booktitle}{\emph{Proceedings of the ACM Turing Award Celebration Conference-China 2024}}. \bibinfo{pages}{187--192}.
\newblock


\bibitem[Chen et~al\mbox{.}(2025)]%
        {chen2025shieldagent}
\bibfield{author}{\bibinfo{person}{Zhaorun Chen}, \bibinfo{person}{Mintong Kang}, {and} \bibinfo{person}{Bo Li}.} \bibinfo{year}{2025}\natexlab{}.
\newblock \showarticletitle{SHIELDAGENT: Shielding Agents via Verifiable Safety Policy Reasoning}.
\newblock \bibinfo{journal}{\emph{Proceedings of Machine Learning Research}}  \bibinfo{volume}{267} (\bibinfo{year}{2025}), \bibinfo{pages}{8313--8344}.
\newblock


\bibitem[Fan and Min(2025)]%
        {fan2025web3agent}
\bibfield{author}{\bibinfo{person}{Sizheng Fan} {and} \bibinfo{person}{Tian Min}.} \bibinfo{year}{2025}\natexlab{}.
\newblock \showarticletitle{Web3Agent: Automating On-Chain Operations via Natural Language Interfaces}.
\newblock \bibinfo{journal}{\emph{ACM Transactions on the Web}} (\bibinfo{year}{2025}).
\newblock


\bibitem[Fei et~al\mbox{.}(2025)]%
        {fei2025mcp}
\bibfield{author}{\bibinfo{person}{Xiang Fei}, \bibinfo{person}{Xiawu Zheng}, {and} \bibinfo{person}{Hao Feng}.} \bibinfo{year}{2025}\natexlab{}.
\newblock \showarticletitle{Mcp-zero: Active tool discovery for autonomous llm agents}.
\newblock \bibinfo{journal}{\emph{arXiv preprint arXiv:2506.01056}} (\bibinfo{year}{2025}).
\newblock


\bibitem[Ferreira~Torres et~al\mbox{.}(2024)]%
        {torres2024rolling}
\bibfield{author}{\bibinfo{person}{Christof Ferreira~Torres}, \bibinfo{person}{Albin Mamuti}, \bibinfo{person}{Ben Weintraub}, \bibinfo{person}{Cristina Nita-Rotaru}, {and} \bibinfo{person}{Shweta Shinde}.} \bibinfo{year}{2024}\natexlab{}.
\newblock \showarticletitle{Rolling in the shadows: Analyzing the extraction of mev across layer-2 rollups}. In \bibinfo{booktitle}{\emph{Proceedings of the 2024 on ACM SIGSAC Conference on Computer and Communications Security}}. \bibinfo{pages}{2591--2605}.
\newblock


\bibitem[Hou et~al\mbox{.}(2025)]%
        {hou2025model}
\bibfield{author}{\bibinfo{person}{Xinyi Hou}, \bibinfo{person}{Yanjie Zhao}, \bibinfo{person}{Shenao Wang}, {and} \bibinfo{person}{Haoyu Wang}.} \bibinfo{year}{2025}\natexlab{}.
\newblock \showarticletitle{Model context protocol (mcp): Landscape, security threats, and future research directions}.
\newblock \bibinfo{journal}{\emph{ACM Transactions on Software Engineering and Methodology}} (\bibinfo{year}{2025}).
\newblock


\bibitem[Hu et~al\mbox{.}(2025)]%
        {hu2025walletprobe}
\bibfield{author}{\bibinfo{person}{Xiaohui Hu}, \bibinfo{person}{Ningyu He}, {and} \bibinfo{person}{Haoyu Wang}.} \bibinfo{year}{2025}\natexlab{}.
\newblock \showarticletitle{WalletProbe: A Testing Framework for Browser-based Cryptocurrency Wallet Extensions}.
\newblock \bibinfo{journal}{\emph{arXiv preprint arXiv:2504.11735}} (\bibinfo{year}{2025}).
\newblock


\bibitem[Jimenez et~al\mbox{.}(2024)]%
        {jimenez2024swe}
\bibfield{author}{\bibinfo{person}{Carlos~E Jimenez}, \bibinfo{person}{John Yang}, \bibinfo{person}{Alexander Wettig}, \bibinfo{person}{Shunyu Yao}, \bibinfo{person}{Kexin Pei}, \bibinfo{person}{Ofir Press}, {and} \bibinfo{person}{Karthik Narasimhan}.} \bibinfo{year}{2024}\natexlab{}.
\newblock \showarticletitle{Swe-bench: Can language models resolve real-world github issues?}. In \bibinfo{booktitle}{\emph{International Conference on Learning Representations}}, Vol.~\bibinfo{volume}{2024}. \bibinfo{pages}{54107--54157}.
\newblock


\bibitem[Jin et~al\mbox{.}(2024)]%
        {jin2024decoagent}
\bibfield{author}{\bibinfo{person}{Anan Jin}, \bibinfo{person}{Yuhang Ye}, \bibinfo{person}{Brian Lee}, {and} \bibinfo{person}{Yuansong Qiao}.} \bibinfo{year}{2024}\natexlab{}.
\newblock \showarticletitle{Decoagent: Large language model empowered decentralized autonomous collaboration agents based on smart contracts}.
\newblock \bibinfo{journal}{\emph{IEEe Access}}  \bibinfo{volume}{12} (\bibinfo{year}{2024}), \bibinfo{pages}{155234--155245}.
\newblock


\bibitem[Jin et~al\mbox{.}(2025)]%
        {jin2025search}
\bibfield{author}{\bibinfo{person}{Bowen Jin}, \bibinfo{person}{Hansi Zeng}, \bibinfo{person}{Zhenrui Yue}, \bibinfo{person}{Jinsung Yoon}, \bibinfo{person}{Sercan Arik}, \bibinfo{person}{Dong Wang}, \bibinfo{person}{Hamed Zamani}, {and} \bibinfo{person}{Jiawei Han}.} \bibinfo{year}{2025}\natexlab{}.
\newblock \showarticletitle{Search-r1: Training llms to reason and leverage search engines with reinforcement learning}.
\newblock \bibinfo{journal}{\emph{arXiv preprint arXiv:2503.09516}} (\bibinfo{year}{2025}).
\newblock


\bibitem[Li et~al\mbox{.}(2024)]%
        {li2024cryptotrade}
\bibfield{author}{\bibinfo{person}{Yuan Li}, \bibinfo{person}{Bingqiao Luo}, \bibinfo{person}{Qian Wang}, \bibinfo{person}{Nuo Chen}, \bibinfo{person}{Xu Liu}, {and} \bibinfo{person}{Bingsheng He}.} \bibinfo{year}{2024}\natexlab{}.
\newblock \showarticletitle{CryptoTrade: A reflective LLM-based agent to guide zero-shot cryptocurrency trading}. In \bibinfo{booktitle}{\emph{Proceedings of the 2024 Conference on Empirical Methods in Natural Language Processing}}. \bibinfo{pages}{1094--1106}.
\newblock


\bibitem[Liu et~al\mbox{.}(2025)]%
        {liu2025blockchain}
\bibfield{author}{\bibinfo{person}{Detian Liu}, \bibinfo{person}{Jianbiao Zhang}, \bibinfo{person}{Yifan Wang}, \bibinfo{person}{Hong Shen}, \bibinfo{person}{Zhaoqian Zhang}, {and} \bibinfo{person}{Tao Ye}.} \bibinfo{year}{2025}\natexlab{}.
\newblock \showarticletitle{Blockchain smart contract security: Threats and mitigation strategies in a lifecycle perspective}.
\newblock \bibinfo{journal}{\emph{Comput. Surveys}} \bibinfo{volume}{58}, \bibinfo{number}{4} (\bibinfo{year}{2025}), \bibinfo{pages}{1--34}.
\newblock


\bibitem[Luo et~al\mbox{.}(2025)]%
        {luo2025agrail}
\bibfield{author}{\bibinfo{person}{Weidi Luo}, \bibinfo{person}{Shenghong Dai}, \bibinfo{person}{Xiaogeng Liu}, \bibinfo{person}{Suman Banerjee}, \bibinfo{person}{Huan Sun}, \bibinfo{person}{Muhao Chen}, {and} \bibinfo{person}{Chaowei Xiao}.} \bibinfo{year}{2025}\natexlab{}.
\newblock \showarticletitle{Agrail: A lifelong agent guardrail with effective and adaptive safety detection}. In \bibinfo{booktitle}{\emph{Proceedings of the 63rd Annual Meeting of the Association for Computational Linguistics (Volume 1: Long Papers)}}. \bibinfo{pages}{8104--8139}.
\newblock


\bibitem[Madaan et~al\mbox{.}(2023)]%
        {madaan2023self}
\bibfield{author}{\bibinfo{person}{Aman Madaan}, \bibinfo{person}{Niket Tandon}, \bibinfo{person}{Prakhar Gupta}, \bibinfo{person}{Skyler Hallinan}, \bibinfo{person}{Luyu Gao}, \bibinfo{person}{Sarah Wiegreffe}, \bibinfo{person}{Uri Alon}, \bibinfo{person}{Nouha Dziri}, \bibinfo{person}{Shrimai Prabhumoye}, \bibinfo{person}{Yiming Yang}, {et~al\mbox{.}}} \bibinfo{year}{2023}\natexlab{}.
\newblock \showarticletitle{Self-refine: Iterative refinement with self-feedback}.
\newblock \bibinfo{journal}{\emph{Advances in neural information processing systems}}  \bibinfo{volume}{36} (\bibinfo{year}{2023}), \bibinfo{pages}{46534--46594}.
\newblock


\bibitem[Patil et~al\mbox{.}(2024)]%
        {patil2024gorilla}
\bibfield{author}{\bibinfo{person}{Shishir~G Patil}, \bibinfo{person}{Tianjun Zhang}, \bibinfo{person}{Xin Wang}, {and} \bibinfo{person}{Joseph~E Gonzalez}.} \bibinfo{year}{2024}\natexlab{}.
\newblock \showarticletitle{Gorilla: Large language model connected with massive apis}.
\newblock \bibinfo{journal}{\emph{Advances in Neural Information Processing Systems}}  \bibinfo{volume}{37} (\bibinfo{year}{2024}), \bibinfo{pages}{126544--126565}.
\newblock


\bibitem[Qi et~al\mbox{.}(2026)]%
        {qi2026towards}
\bibfield{author}{\bibinfo{person}{Minfeng Qi}, \bibinfo{person}{Tianqing Zhu}, \bibinfo{person}{Lefeng Zhang}, \bibinfo{person}{Ningran Li}, \bibinfo{person}{Yu-an Tan}, {and} \bibinfo{person}{Wanlei Zhou}.} \bibinfo{year}{2026}\natexlab{}.
\newblock \showarticletitle{Towards transparent and incentive-compatible collaboration in decentralized llm multi-agent systems: A blockchain-driven approach}.
\newblock \bibinfo{journal}{\emph{IEEE Transactions on Network Science and Engineering}} (\bibinfo{year}{2026}).
\newblock


\bibitem[Qian et~al\mbox{.}(2025)]%
        {qian2025agents}
\bibfield{author}{\bibinfo{person}{Lingfei Qian}, \bibinfo{person}{Xueqing Peng}, \bibinfo{person}{Yan Wang}, \bibinfo{person}{Vincent~Jim Zhang}, \bibinfo{person}{Huan He}, \bibinfo{person}{Hanley Smith}, \bibinfo{person}{Yi Han}, \bibinfo{person}{Yueru He}, \bibinfo{person}{Haohang Li}, \bibinfo{person}{Yupeng Cao}, {et~al\mbox{.}}} \bibinfo{year}{2025}\natexlab{}.
\newblock \showarticletitle{When agents trade: Live multi-market trading benchmark for llm agents}.
\newblock \bibinfo{journal}{\emph{arXiv preprint arXiv:2510.11695}} (\bibinfo{year}{2025}).
\newblock


\bibitem[Qin et~al\mbox{.}(2024)]%
        {qin2024toolllm}
\bibfield{author}{\bibinfo{person}{Yujia Qin}, \bibinfo{person}{Shihao Liang}, \bibinfo{person}{Yining Ye}, \bibinfo{person}{Kunlun Zhu}, \bibinfo{person}{Lan Yan}, \bibinfo{person}{Yaxi Lu}, \bibinfo{person}{Yankai Lin}, \bibinfo{person}{Xin Cong}, \bibinfo{person}{Xiangru Tang}, \bibinfo{person}{Bill Qian}, {et~al\mbox{.}}} \bibinfo{year}{2024}\natexlab{}.
\newblock \showarticletitle{Toolllm: Facilitating large language models to master 16000+ real-world apis}. In \bibinfo{booktitle}{\emph{International Conference on Learning Representations}}, Vol.~\bibinfo{volume}{2024}. \bibinfo{pages}{9695--9717}.
\newblock


\bibitem[Ray(2025)]%
        {ray2025survey}
\bibfield{author}{\bibinfo{person}{Partha~Pratim Ray}.} \bibinfo{year}{2025}\natexlab{}.
\newblock \showarticletitle{A survey on model context protocol: Architecture, state-of-the-art, challenges and future directions}.
\newblock \bibinfo{journal}{\emph{Authorea Preprints}} (\bibinfo{year}{2025}).
\newblock


\bibitem[Schick et~al\mbox{.}(2023)]%
        {schick2023toolformer}
\bibfield{author}{\bibinfo{person}{Timo Schick}, \bibinfo{person}{Jane Dwivedi-Yu}, \bibinfo{person}{Roberto Dess{\`\i}}, \bibinfo{person}{Roberta Raileanu}, \bibinfo{person}{Maria Lomeli}, \bibinfo{person}{Eric Hambro}, \bibinfo{person}{Luke Zettlemoyer}, \bibinfo{person}{Nicola Cancedda}, {and} \bibinfo{person}{Thomas Scialom}.} \bibinfo{year}{2023}\natexlab{}.
\newblock \showarticletitle{Toolformer: Language models can teach themselves to use tools}.
\newblock \bibinfo{journal}{\emph{Advances in neural information processing systems}}  \bibinfo{volume}{36} (\bibinfo{year}{2023}), \bibinfo{pages}{68539--68551}.
\newblock


\bibitem[Shen et~al\mbox{.}(2023)]%
        {shen2023hugginggpt}
\bibfield{author}{\bibinfo{person}{Yongliang Shen}, \bibinfo{person}{Kaitao Song}, \bibinfo{person}{Xu Tan}, \bibinfo{person}{Dongsheng Li}, \bibinfo{person}{Weiming Lu}, {and} \bibinfo{person}{Yueting Zhuang}.} \bibinfo{year}{2023}\natexlab{}.
\newblock \showarticletitle{Hugginggpt: Solving ai tasks with chatgpt and its friends in hugging face}.
\newblock \bibinfo{journal}{\emph{Advances in Neural Information Processing Systems}}  \bibinfo{volume}{36} (\bibinfo{year}{2023}), \bibinfo{pages}{38154--38180}.
\newblock


\bibitem[Shinn et~al\mbox{.}(2023)]%
        {shinn2023reflexion}
\bibfield{author}{\bibinfo{person}{Noah Shinn}, \bibinfo{person}{Federico Cassano}, \bibinfo{person}{Ashwin Gopinath}, \bibinfo{person}{Karthik Narasimhan}, {and} \bibinfo{person}{Shunyu Yao}.} \bibinfo{year}{2023}\natexlab{}.
\newblock \showarticletitle{Reflexion: Language agents with verbal reinforcement learning}.
\newblock \bibinfo{journal}{\emph{Advances in neural information processing systems}}  \bibinfo{volume}{36} (\bibinfo{year}{2023}), \bibinfo{pages}{8634--8652}.
\newblock


\bibitem[Steinberger(2025)]%
        {openclaw2025}
\bibfield{author}{\bibinfo{person}{Peter Steinberger}.} \bibinfo{year}{2025}\natexlab{}.
\newblock \bibinfo{booktitle}{\emph{OpenClaw}}.
\newblock
\urldef\tempurl%
\url{https://github.com/openclaw/openclaw}
\showURL{%
\tempurl}


\bibitem[Sun et~al\mbox{.}(2024)]%
        {sun2024gptscan}
\bibfield{author}{\bibinfo{person}{Yuqiang Sun}, \bibinfo{person}{Daoyuan Wu}, \bibinfo{person}{Yue Xue}, \bibinfo{person}{Han Liu}, \bibinfo{person}{Haijun Wang}, \bibinfo{person}{Zhengzi Xu}, \bibinfo{person}{Xiaofei Xie}, {and} \bibinfo{person}{Yang Liu}.} \bibinfo{year}{2024}\natexlab{}.
\newblock \showarticletitle{Gptscan: Detecting logic vulnerabilities in smart contracts by combining gpt with program analysis}. In \bibinfo{booktitle}{\emph{Proceedings of the IEEE/ACM 46th international conference on software engineering}}. \bibinfo{pages}{1--13}.
\newblock


\bibitem[Wang et~al\mbox{.}(2025)]%
        {wang2025agentspec}
\bibfield{author}{\bibinfo{person}{Haoyu Wang}, \bibinfo{person}{Christopher~M Poskitt}, {and} \bibinfo{person}{Jun Sun}.} \bibinfo{year}{2025}\natexlab{}.
\newblock \showarticletitle{Agentspec: Customizable runtime enforcement for safe and reliable llm agents}.
\newblock \bibinfo{journal}{\emph{arXiv preprint arXiv:2503.18666}} (\bibinfo{year}{2025}).
\newblock


\bibitem[Wang et~al\mbox{.}(2024a)]%
        {wang2024survey}
\bibfield{author}{\bibinfo{person}{Lei Wang}, \bibinfo{person}{Chen Ma}, \bibinfo{person}{Xueyang Feng}, \bibinfo{person}{Zeyu Zhang}, \bibinfo{person}{Hao Yang}, \bibinfo{person}{Jingsen Zhang}, \bibinfo{person}{Zhiyuan Chen}, \bibinfo{person}{Jiakai Tang}, \bibinfo{person}{Xu Chen}, \bibinfo{person}{Yankai Lin}, {et~al\mbox{.}}} \bibinfo{year}{2024}\natexlab{a}.
\newblock \showarticletitle{A survey on large language model based autonomous agents}.
\newblock \bibinfo{journal}{\emph{Frontiers of Computer Science}} \bibinfo{volume}{18}, \bibinfo{number}{6} (\bibinfo{year}{2024}), \bibinfo{pages}{186345}.
\newblock


\bibitem[Wang et~al\mbox{.}(2024b)]%
        {wang2024smartinv}
\bibfield{author}{\bibinfo{person}{Sally~Junsong Wang}, \bibinfo{person}{Kexin Pei}, {and} \bibinfo{person}{Junfeng Yang}.} \bibinfo{year}{2024}\natexlab{b}.
\newblock \showarticletitle{Smartinv: Multimodal learning for smart contract invariant inference}. In \bibinfo{booktitle}{\emph{2024 IEEE Symposium on Security and Privacy (SP)}}. IEEE, \bibinfo{pages}{2217--2235}.
\newblock


\bibitem[Xi et~al\mbox{.}(2024)]%
        {xi2024pomabuster}
\bibfield{author}{\bibinfo{person}{Rui Xi}, \bibinfo{person}{Zehua Wang}, {and} \bibinfo{person}{Karthik Pattabiraman}.} \bibinfo{year}{2024}\natexlab{}.
\newblock \showarticletitle{Pomabuster: Detecting price oracle manipulation attacks in decentralized finance}. In \bibinfo{booktitle}{\emph{2024 IEEE Symposium on Security and Privacy (SP)}}. IEEE, \bibinfo{pages}{3923--3942}.
\newblock


\bibitem[Xi et~al\mbox{.}(2025)]%
        {xi2025rise}
\bibfield{author}{\bibinfo{person}{Zhiheng Xi}, \bibinfo{person}{Wenxiang Chen}, \bibinfo{person}{Xin Guo}, \bibinfo{person}{Wei He}, \bibinfo{person}{Yiwen Ding}, \bibinfo{person}{Boyang Hong}, \bibinfo{person}{Ming Zhang}, \bibinfo{person}{Junzhe Wang}, \bibinfo{person}{Senjie Jin}, \bibinfo{person}{Enyu Zhou}, {et~al\mbox{.}}} \bibinfo{year}{2025}\natexlab{}.
\newblock \showarticletitle{The rise and potential of large language model based agents: A survey}.
\newblock \bibinfo{journal}{\emph{Science China Information Sciences}} \bibinfo{volume}{68}, \bibinfo{number}{2} (\bibinfo{year}{2025}), \bibinfo{pages}{121101}.
\newblock


\bibitem[Xiang et~al\mbox{.}(2024)]%
        {xiang2024guardagent}
\bibfield{author}{\bibinfo{person}{Zhen Xiang}, \bibinfo{person}{Linzhi Zheng}, \bibinfo{person}{Yanjie Li}, \bibinfo{person}{Junyuan Hong}, \bibinfo{person}{Qinbin Li}, \bibinfo{person}{Han Xie}, \bibinfo{person}{Jiawei Zhang}, \bibinfo{person}{Zidi Xiong}, \bibinfo{person}{Chulin Xie}, \bibinfo{person}{Carl Yang}, {et~al\mbox{.}}} \bibinfo{year}{2024}\natexlab{}.
\newblock \showarticletitle{Guardagent: Safeguard llm agents by a guard agent via knowledge-enabled reasoning}.
\newblock \bibinfo{journal}{\emph{arXiv preprint arXiv:2406.09187}} (\bibinfo{year}{2024}).
\newblock


\bibitem[Xu et~al\mbox{.}(2025)]%
        {xu2025auto}
\bibfield{author}{\bibinfo{person}{Jiahua Xu}, \bibinfo{person}{Yebo Feng}, \bibinfo{person}{Daniel Perez}, {and} \bibinfo{person}{Benjamin Livshits}.} \bibinfo{year}{2025}\natexlab{}.
\newblock \showarticletitle{Auto. gov: learning-based governance for decentralized finance (DeFi)}.
\newblock \bibinfo{journal}{\emph{IEEE Transactions on Services Computing}} (\bibinfo{year}{2025}).
\newblock


\bibitem[Xu et~al\mbox{.}(2026)]%
        {xu2026mem}
\bibfield{author}{\bibinfo{person}{Wujiang Xu}, \bibinfo{person}{Zujie Liang}, \bibinfo{person}{Kai Mei}, \bibinfo{person}{Hang Gao}, \bibinfo{person}{Juntao Tan}, {and} \bibinfo{person}{Yongfeng Zhang}.} \bibinfo{year}{2026}\natexlab{}.
\newblock \showarticletitle{A-mem: Agentic memory for llm agents}.
\newblock \bibinfo{journal}{\emph{Advances in Neural Information Processing Systems}}  \bibinfo{volume}{38} (\bibinfo{year}{2026}), \bibinfo{pages}{17577--17604}.
\newblock


\bibitem[Yang et~al\mbox{.}(2024)]%
        {yang2024swe}
\bibfield{author}{\bibinfo{person}{John Yang}, \bibinfo{person}{Carlos Jimenez}, \bibinfo{person}{Alexander Wettig}, \bibinfo{person}{Kilian Lieret}, \bibinfo{person}{Shunyu Yao}, \bibinfo{person}{Karthik Narasimhan}, {and} \bibinfo{person}{Ofir Press}.} \bibinfo{year}{2024}\natexlab{}.
\newblock \showarticletitle{Swe-agent: Agent-computer interfaces enable automated software engineering}.
\newblock \bibinfo{journal}{\emph{Advances in Neural Information Processing Systems}}  \bibinfo{volume}{37} (\bibinfo{year}{2024}), \bibinfo{pages}{50528--50652}.
\newblock


\bibitem[Yao et~al\mbox{.}({[n.\,d.]})]%
        {yao2022react}
\bibfield{author}{\bibinfo{person}{Shunyu Yao}, \bibinfo{person}{Jeffrey Zhao}, \bibinfo{person}{Dian Yu}, \bibinfo{person}{Izhak Shafran}, \bibinfo{person}{Karthik~R Narasimhan}, {and} \bibinfo{person}{Yuan Cao}.} \bibinfo{year}{[n.\,d.]}\natexlab{}.
\newblock \showarticletitle{ReAct: Synergizing Reasoning and Acting in Language Models}. In \bibinfo{booktitle}{\emph{NeurIPS 2022 Foundation Models for Decision Making Workshop}}.
\newblock


\bibitem[Zhang et~al\mbox{.}(2025a)]%
        {zhang2025following}
\bibfield{author}{\bibinfo{person}{Bosi Zhang}, \bibinfo{person}{Ningyu He}, \bibinfo{person}{Xiaohui Hu}, \bibinfo{person}{Kai Ma}, {and} \bibinfo{person}{Haoyu Wang}.} \bibinfo{year}{2025}\natexlab{a}.
\newblock \showarticletitle{Following Devils' Footprint: Towards Real-time Detection of Price Manipulation Attacks}. In \bibinfo{booktitle}{\emph{34th USENIX Security Symposium (USENIX Security 25)}}. \bibinfo{pages}{4127--4145}.
\newblock


\bibitem[Zhang et~al\mbox{.}(2025b)]%
        {zhang2025crossguard}
\bibfield{author}{\bibinfo{person}{Xu Zhang}, \bibinfo{person}{Hao Li}, {and} \bibinfo{person}{Zhichao Lu}.} \bibinfo{year}{2025}\natexlab{b}.
\newblock \showarticletitle{CrossGuard: Safeguarding MLLMs against Joint-Modal Implicit Malicious Attacks}.
\newblock \bibinfo{journal}{\emph{arXiv preprint arXiv:2510.17687}} (\bibinfo{year}{2025}).
\newblock


\bibitem[Zhou et~al\mbox{.}(2023)]%
        {zhou2023sok}
\bibfield{author}{\bibinfo{person}{Liyi Zhou}, \bibinfo{person}{Xihan Xiong}, \bibinfo{person}{Jens Ernstberger}, \bibinfo{person}{Stefanos Chaliasos}, \bibinfo{person}{Zhipeng Wang}, \bibinfo{person}{Ye Wang}, \bibinfo{person}{Kaihua Qin}, \bibinfo{person}{Roger Wattenhofer}, \bibinfo{person}{Dawn Song}, {and} \bibinfo{person}{Arthur Gervais}.} \bibinfo{year}{2023}\natexlab{}.
\newblock \showarticletitle{Sok: Decentralized finance (defi) attacks}. In \bibinfo{booktitle}{\emph{2023 IEEE Symposium on Security and Privacy (SP)}}. IEEE, \bibinfo{pages}{2444--2461}.
\newblock


\bibitem[Zhou et~al\mbox{.}(2024)]%
        {zhou2024webarena}
\bibfield{author}{\bibinfo{person}{Shuyan Zhou}, \bibinfo{person}{Frank~F Xu}, \bibinfo{person}{Hao Zhu}, \bibinfo{person}{Xuhui Zhou}, \bibinfo{person}{Robert Lo}, \bibinfo{person}{Abishek Sridhar}, \bibinfo{person}{Xianyi Cheng}, \bibinfo{person}{Tianyue Ou}, \bibinfo{person}{Yonatan Bisk}, \bibinfo{person}{Daniel Fried}, {et~al\mbox{.}}} \bibinfo{year}{2024}\natexlab{}.
\newblock \showarticletitle{Webarena: A realistic web environment for building autonomous agents}. In \bibinfo{booktitle}{\emph{International Conference on Learning Representations}}, Vol.~\bibinfo{volume}{2024}. \bibinfo{pages}{15585--15606}.
\newblock


\end{thebibliography}

\appendix









\end{document}